\documentclass[journal]{IEEEtran}
\usepackage{cite}
\usepackage{amsmath,amssymb}
\usepackage{multirow}
\usepackage{graphicx}

\usepackage{color}

\begin{document}
\title{Multiple Visual-Semantic Embedding for Video Retrieval from Query Sentence}
\author{
  Huy~Manh~Nguyen,
  Tomo~Miyazaki,
  Yoshihiro~Sugaya,
  and~Shinichiro~Omachi,
  \thanks{H. M. Nguyen was with the Department of Communication Engineering at Tohoku University, Sendai, Miyagi JAPAN 9808579}%
\thanks{T. Miyazaki, Y. Sugaya, and S. Omachi are with Tohoku University. e-mail: (see https://tomomiyazaki.github.io).}
}


\maketitle

\begin{abstract}
Visual-semantic embedding aims to learn a joint embedding space where related video and sentence instances are located close to each other. Most existing methods put instances in a single embedding space. However, they struggle to embed instances due to the difficulty of matching visual dynamics in videos to textual features in sentences. A single space is not enough to accommodate various videos and sentences.
In this paper, we propose a novel framework that maps instances into multiple individual embedding spaces so that we can capture multiple relationships between instances, leading to compelling video retrieval. We propose to produce a final similarity between instances by fusing similarities measured in each embedding space using a weighted sum strategy. We determine the weights according to a sentence. Therefore, we can flexibly emphasize an embedding space.
We conducted sentence-to-video retrieval experiments on a benchmark dataset. The proposed method achieved superior performance, and the results are competitive to state-of-the-art methods. These experimental results demonstrated the effectiveness of the proposed multiple embedding approach compared to existing methods.
\end{abstract}


%

\section{Introduction}
\IEEEPARstart{V}{ideo} has become an essential source for humans to learn and acquire knowledge. Due to the increased demand for sharing and accumulating information, there is a massive amount of video being produced out there in the world every day. However, compared to images, videos usually contain much semantic information, and thus it is hard for humans to organize videos. Therefore, it is critical to developing an algorithm that can efficiently perform multimedia analysis and automatically understand the semantic information of videos.

A common approach for video analysis and understanding is to form a joint embedding space of video and sentence using multimodal learning. Similar to the fact that humans experience the world with multiple senses, the goal of multimodal learning is to develop a model that can simultaneously process multiple modalities, such as visual, text, and audio, in an integrated manner by constructing a joint embedding space. Such models can map various modalities into a shared Euclidean space where distances and directions capture useful semantic relationships. It enables us to not only learn the semantic presentations of videos by leveraging other modalities information but also show great potential in tasks such as cross-modal retrieval and visual recognition.

Recent works in learning an embedding space bridge the gap between sentence and visual information by utilizing advancements in image and language understanding~\cite{bib:NIPS:2013:devise, Engilberge2018FindingBI}. Most approaches build the embedding space by connecting visual and textual embedding paths. Generally, a visual path uses a Convolutional Neural Network (CNN) to transform visual appearances into a vector. Likewise, Recurrent Neural Network (RNN) embeds sentences in a textual path. However, capturing the relationship between video and sentence remains challenging. Recent works suffer from extracting visual dynamics in a video~\cite{Otani2016LearningJR, bib:ICCV:2017:Hend, bib:ICCV:2017:tall, bib:AAAI:2019:Xu}.

In this paper, we propose a novel framework equipped with multiple embedding networks so that we can capture various relationships between video and sentence, leading to more compelling video retrieval.
Precisely, one network captures the relationship between an overall appearance in the video and a textual feature. Others consider consecutive appearances or action features. Thus, the networks learn their own embedding spaces. We fuse the similarities measured in the multiple spaces using the weighted summing strategy to produce the final similarity between video and sentence.

The main contribution of this paper is a novel approach to measure the similarity between video and sentence by fusing similarities in multiple embedding spaces. Consequently, we can measure the similarity with multiple understandings and relationships of video and sentence. Besides, we emphasize that the proposed method can quickly expand the number of embedding spaces. We demonstrated the effectiveness of the proposed method by the experimental results. We conducted video retrieval experiments using query sentences on the standard benchmark dataset and demonstrated an improvement of our approach compared to existing methods.

\section{Related Work}

\subsection*{Vision and Language Understanding}
There have been many efforts in connecting vision and language, which focus on building a joint visual-semantic space~\cite{bib:TPAMI:2017:Karpathy}. Various applications in the area of computer vision needs such a joint space to realize tagging~\cite{bib:NIPS:2013:devise}, retrieval~\cite{bib:IJCV:2014:Gong}, captioning~\cite{bib:IJCV:2014:Gong}, and visual question answering~\cite{jang-CVPR-2017}.

Recent works in image-to-text retrieval embed image and sentence into a joint embedding trained with ranking loss. A penalty is applied when an incorrect sentence is ranked higher than the correct sentence~\cite{bib:arxiv:2016:Wang, bib:ICCV:2019:Saliency, bib:CVPR:2019:Wu:Unified, bib:CVPR:2018:Gu:Look, bib:OrderLearning:TPAMI:2020}. Another popular loss function is triplet ranking~\cite{bib:NIPS:2013:devise, bib:arxiv:2014:Unifying, bib:CVPR:2016:Learning, bib:ECCV:2018:advise}. VSE++ model improves the triplet ranking loss by focusing on the hardest negative samples (the most similar yet incorrect sample in a mini-batch)~\cite{Bib:BMVC:2018:VSE++}.

\subsection*{Video and Sentence Embedding}
As same as image-text retrieval approaches, most video-to-text retrieval methods learn a joint embedding space~\cite{bib:MAN:CVPR:2019, bib:CVPR:2019:Tsai, bib:CVPR:2019:Song}. The method~\cite{Otani2016LearningJR} incorporates web images searched with a sentence into an embedding process to take into account fine-grained visual concepts. However, the method treats video frames as a set of unrelated images and average them out to get a final video embedding vector. Thus, it may lead to inefficiency in learning an embedding space since temporal information of the video is lost.

Mithun et al. tackle this issue~\cite{Mithun2018JointEW} by learning two different embedding spaces to consider temporal and appearance. Also, they extract audio features from the video for learning space. This approach achieved accurate performance for the sentence-to-video retrieval task. However, this approach puts equal importance on both embedding spaces. Practically, equal importance does not work well. There are cases that one embedding space is more important than the others in capturing semantic similarity between video and sentence. Therefore, we propose a novel mechanism emphasizing a space so that we can know the critical visual cues.

\section{The Proposed Method}

\subsection{Overview} \label{sec:overview}
We describe the problem of learning a joint embedding space for video and sentence embedding.
Given video and sentence instances which are sequences of frames and words, the aim is to train embedding functions that map them into a joint embedding space. Formally, we use embedding functions ${\rm f} : \mathcal{X} \to \mathcal{Z}$ and ${\rm g} : \mathcal{Y} \to \mathcal{Z}$, where $\mathcal{X}$ and $\mathcal{Y}$ are video and sentence domains, respectively. $\mathcal{Z}$ is a joint embedding space. The similarity ${\rm s}( {\rm f}(x),{\rm g}(y))$ between $\mathcal{X}$ and $\mathcal{Y}$ are calculated in $\mathcal{Z}$ by a certain measurement. Therefore, the ultimate goal is learning ${ \rm f}$ and ${\rm g}$ satisfying the following equation: ${\rm s}( {\rm f}(x_i), {\rm g}(y_i) ) > {\rm s}( {\rm f}(x_i), {\rm g}(y_j) ), \forall \ i \neq j.$ This encourages similarity will increase in a same pair $x_i$ and $y_i$, whereas it decreases in a different pair $x_i$ and $y_j$.

\begin{figure*}[!t] \centering
  \includegraphics[width=7in]{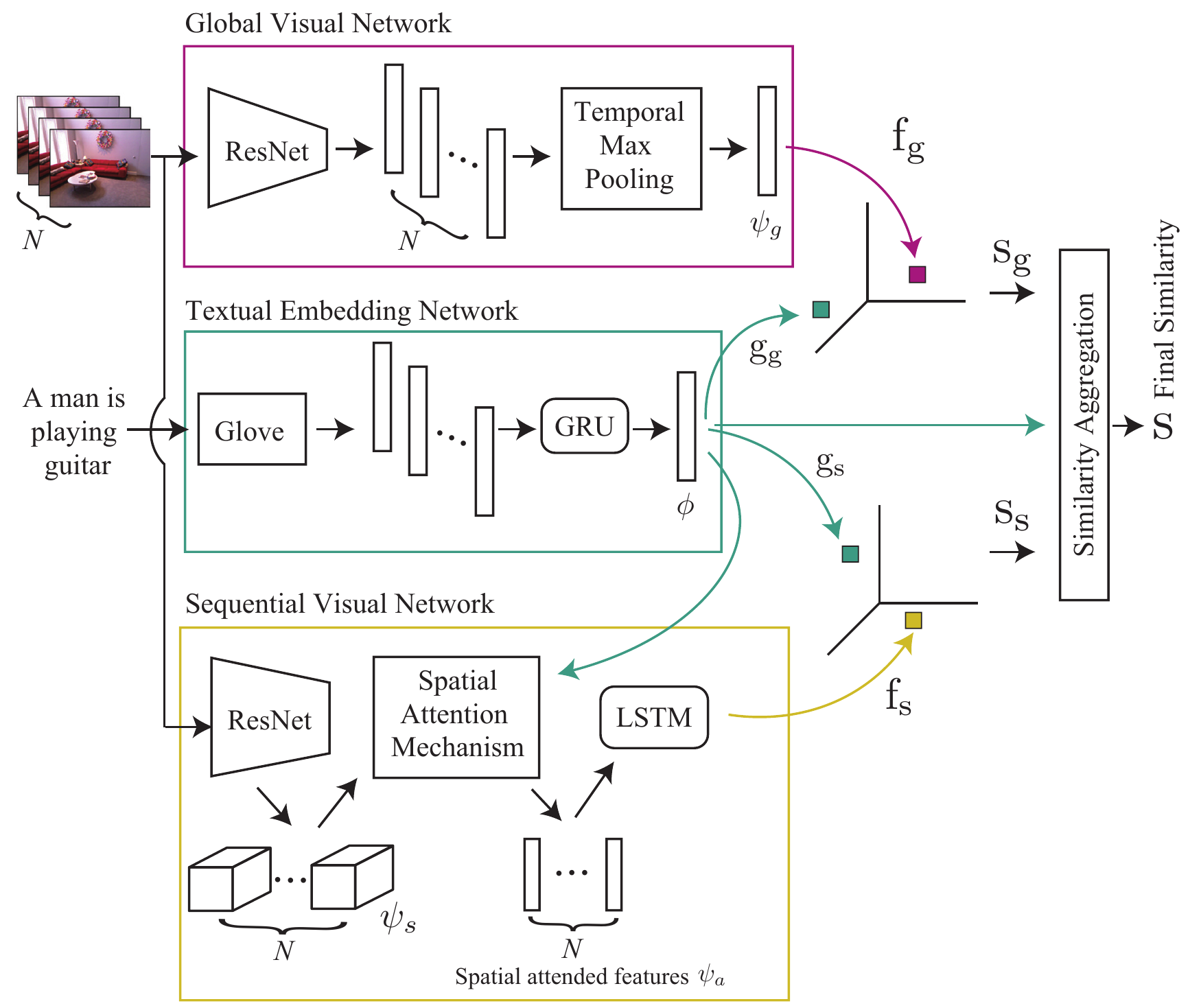}
  \caption{Overview of the proposed framework. $\rm f_g$, $\rm f_s$, $\rm g_g$, and $g_s$ are embedding functions.} \label{fig:overview}
\end{figure*}

As we illustrated the overview of the proposed framework in Fig.~\ref{fig:overview}, there are two networks for embedding videos: the global and the sequential visual networks. These networks have their counterparts that are embedding the sentence. Thus, we develop multiple visual and textual networks that are responsible for embedding a video and a sentence, respectively. We form one embedding space by merging two networks (one visual and one textual) so that we can align a visual feature to textual information. Specifically, the global visual network aligns average visual information of the video to the sentence. Likewise, the sequential visual network aligns a temporal context to the sentence. Consequently, the networks receive a video and sentence as inputs and map them into the joint embedding spaces.

We propose to fuse similarities measured in the embedding spaces to produce the final similarity. One embedding space has the global visual network extracting a visual object feature from the video. Thus, this embedding space describes a relationship between global visual appearances and textual features. The other embedding space captures the relationship between sequential appearances and textual features. The similarity scores of the two embedding spaces are then combined using the weighted sum to put more emphasis on one embedding space than the other space.

Our motivation is twofold. Firstly, since videos and sentences require different attention, we aim to develop a more robust similarity measurement by combining multiple embedding spaces in a weighted manner instead of using a hard-coded average. Secondly, in order to highlight spatial and temporal information of videos according to a sentence, we need to develop a mechanism that utilizes textual information to emphasize spatial and temporal features.

\subsection{Textual Embedding Network}
We decompose the sentence to variable-length sequences of tokens. Then, we transform each token into a 300-dimensional vector representation by using the pre-trained GloVe model~\cite{pennington-etal-2014-glove}. The length of the tokens depends on the sentence. Therefore, in order to obtain a fixed-length meaningful representation, we encode the GloVe vectors using the Gated Recurrent Unit (GRU)~\cite{Chung2014EmpiricalEO} with $H$ hidden states, resulting in a vector $\phi(y) \in \mathbb{R}^H$. We set $H = 512$. This embedded vector $\phi(y)$ goes to four processing modules: global and sequential visual networks, spatial attention mechanism, and similarity aggregation. We further transform $\phi(y)$ in each processing module.

\subsection{Global Visual Network}
The global visual network aims to obtain a vector representation that is a general visual feature over the video frames. We divide the input video frames into $N$ chunks, and one frame is sampled from each chunk by random sampling. We set $N = 20$. We extract visual features from the sampled frames using the ResNet-152 pre-trained on the ImageNet dataset~\cite{He2015DeepRL}. Specifically, we resize the sampled frames to $224 \times 224$, and then the ResNet encodes them, resulting in 2048-dimensional $N$ vectors. Note that we extract the vectors directly from the last fully connected layer of the ResNet. Subsequently, we apply average-pooling to the $N$ vectors to merge them. Consequently, we obtain a feature vector $\psi_g(x) \in \mathbb{R}^{2048}$ containing a global visual feature in the video.

We learn a joint embedding space of the global visual and textual networks.
As defined in Eq.~\eqref{eq:proposal:fg} and \eqref{eq:proposal:gg}, we use embedding functions to embed $\psi_g(x)$ and $\phi(y)$ into a $D$-dimensional space.
\begin{eqnarray}
  {\rm f_g} (x) &=& W_g \psi_g (x) + b_g \label{eq:proposal:fg}\\
  {\rm g_g} (y) &=& \hat{W}_g \phi(y) + \hat{b}_g \label{eq:proposal:gg}
\end{eqnarray}
There are learnable parameters $W_g \in \mathbb{R}^{2048 \times D}$, $\hat{W}_g \in \mathbb{R}^{H \times D}$, and $b_g, \hat{b}_g \in \mathbb{R}^D$. We set $D = 512$ in this paper.

We use cosine similarity to measure the similarity ${\rm s_g}(y,x)$ between the video and sentence in the joint embedding space.
\begin{equation}
  {\rm s_g}(y,x) = \frac{{\rm f_g}(x) \cdot {\rm g_g}(y)}{\parallel {\rm f_g}(x) \parallel \parallel {\rm g_g}(y)\parallel}
\end{equation}

\subsection{Sequential Visual Network}
Similar to the global visual network, we divide the input video frames into $N$ chunks and take the first frames of each chunk as the input of the sequential visual network. Then, we use the ResNet-152 to extract a $7 \times 7 \times 2048$-dimensional vector $\psi_s(x)$ from each frame at the last convolution layer of the ResNet. The $\psi_s(x)$ contains visual features in spatial regions. Considering spatial regions where we should pay further attention may change by sentences, we need to explore relationships between spatial and textual features. Therefore, we incorporate a spatial attention mechanism into the sequential visual network. We apply the attention mechanism to $\psi_s(x)$ to emphasize spatial regions. Finally, we capture the sequential information of the video using a single layer Long-Short Term Memory (LSTM)~\cite{bib:LSTM} with an $H$-dimensional hidden state. We denote the vector embedded by the sequential visual network as ${\rm f_s}(x) \in \mathbb{R}^D$.

The sequential visual network uses LSTM to capture a meaningful sequential representation of the video. However, the LSTM transforms all spatial details of a video into a flat representation, resulting in losing its spatial reasoning with the sentence. Therefore, we employ the spatial attention mechanism in order to obtain a spatial relationship between video and sentence. Inspired by the work~\cite{jang-CVPR-2017}, we develop the spatial attention mechanism to learn with regions in a frame to attend for each word in the sentence.

\begin{figure}[!t] \centering
  \includegraphics[width=3.5in]{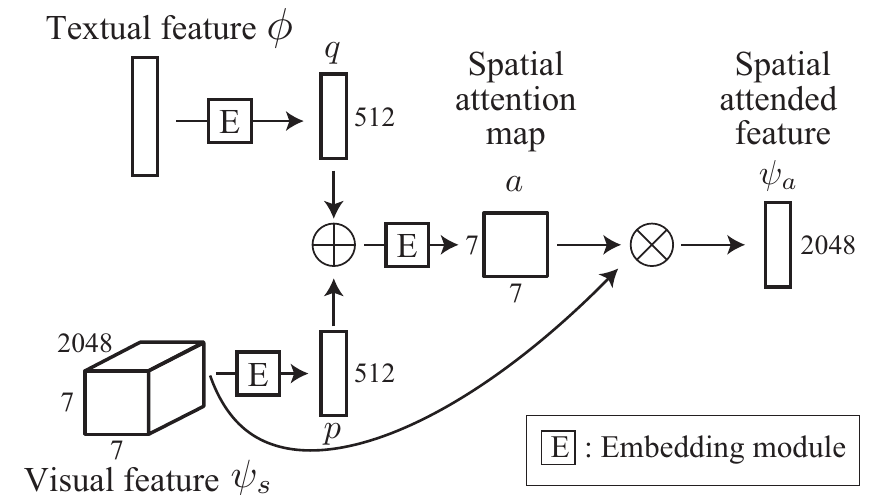}
  \caption{The spatial attention mechanism used in the sequential visual network.} \label{fig:attention}
\end{figure}

We illustrated the spatial attention mechanism in Fig.~\ref{fig:attention}. The mechanism associates the visual feature vector $\psi_s(x)$ with the textual vector $\phi(y)$ to produce a spatial attention map $a \in \mathbb{R}^{7 \times 7}$. Then, we combine $\psi_s(x)$ with $a$ to produce the spatial attended feature $\psi_a(x)$. Formally, $\psi_a(x)$ and $a$ are defined in Eq.~\eqref{eq:attended}~and~\eqref{eq:attention}, respectively.

Specifically, Eq.~\eqref{eq:fs} defines an embedding function of the sequential visual network.
$\gamma$ represents the LSTM, and $\otimes$ is element-wise product.
The $p$ and $q$ mean intermediate outputs, which are 512-dimensional vectors.
We have learnable parameters
$W_a \in \mathbb{R}^{512 \times (7 \times 7)}$,
$W_p \in \mathbb{R}^{(7 \times 7 \times 2048) \times 512}$,
$\hat{W}_q \in \mathbb{R}^{H \times 512}$,
$b_a \in \mathbb{R}^{7 \times 7}$, and 
$b_p, \hat{b}_q \in \mathbb{R}^{512}$.
\begin{eqnarray}
  {\rm f_s}(x) &=& \gamma( \psi_a(x) ) \label{eq:fs}\\
  \psi_a(x) &=& \psi_s(x) \otimes a \label{eq:attended}\\
  a &=& {\rm softmax}( {\rm tanh}( W_a (p + q) + b_a) ) \label{eq:attention}\\
  p &=& {\rm tanh}( W_p \psi_s(x) + b_p)\\
  q &=& {\rm tanh}( \hat{W}_q \phi(y) + \hat{b}_q )
\end{eqnarray}

The joint embedding space of the sequential visual and textual networks is formed by ${\rm f_s}(x)$ and ${\rm g_s}(y)$. We measure the similarity ${\rm s_s}(x,y)$ in this joint embedding space. The formulations are described below, where $\hat{W}_s \in \mathbb{R}^{H \times D}$ and $\hat{b}_s \in \mathbb{R}^D$ are learnable parameters.
\begin{eqnarray}
  {\rm g_s} (y) &=& \hat{W}_s \phi(y) + \hat{b}_s\\
  {\rm s_s}(y,x) &=& \frac{{\rm f_s}(x) \cdot {\rm g_s}(y)}{\parallel {\rm f_s}(x) \parallel \parallel {\rm g_s}(y)\parallel}
\end{eqnarray}

\subsection{Similarity Aggregation}
There are many approaches to aggregate similarities. An average is a straightforward approach. Some cases work well on average. However, the average may cause unexpected behaviors if an inaccurate similarity is considerably high or low. Therefore, we adopt the Mixture of Experts fusion strategy~\cite{Jacobs1991AdaptiveMO} for aggregating similarities with weights that changes according to the input sentence. Consequently, we can emphasize one embedding space using the weights for merging the multiple similarities.

\begin{figure}[t] \centering
  \includegraphics[width=3.5in]{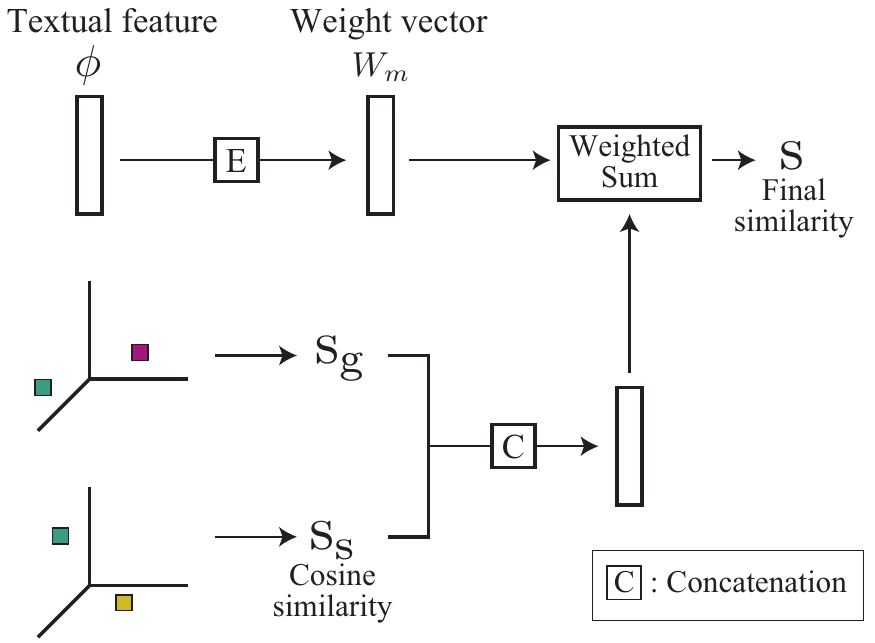}
  \caption{Similarity aggregation} \label{fig:similarity}
\end{figure}

We propose to aggregate the similarities measured in multiple embedding spaces so that we can produce the final similarity ${\rm s}(x,y)$ with various understandings of videos and sentences. We illustrate the proposed similarity aggregation in Fig.~\ref{fig:similarity}. Specifically, we merge the similarities using the weight $W_m \in \mathbb{R}^2$ generated by considering the textual feature. $\hat{W}_t \in \mathbb{R}^{D \times 2}$ is a learnable parameter. ${\rm concat()}$ is a function that concatenates given scalar values.
\begin{eqnarray}
  {\rm s}(x,y) &=& W_m \left ( {\rm concat}( {\rm s_g}(x,y), {\rm s_s}(x,y) ) \right ) \label{eq:sim:dual}\\
  W_m &=& {\rm softmax}( \hat{W}_t \phi(y) ) \label{eq:aggregation:weight}
\end{eqnarray}

\subsection{Optimization}
As described in Section~\ref{sec:overview}, we optimize the proposed architecture by enforcing similarity of a video $x_i$ and its counterpart sentence $y_i$ will be greater than similarities of the video $x_i$ and other sentence $y_j$, such as ${\rm s}(x_i,y_i) \geq {\rm s}(x_i,y_j)$ or ${\rm s}(x_j,y_i)$. We achieve this by using the triplet ranking loss~\cite{Kiros2014UnifyingVE, bib:JMLR:2010:Chechik, bib:ICCV:2007:Frome}, where $\alpha$ is a margin.
\begin{eqnarray}
  \mathcal{L}_{\rm s}( x_i,y_i,y_j ) &=& \max \left \{ 0, \alpha - {\rm s}(x_i,y_i) + {\rm s}(x_i,y_j) \right \}\\
  \mathcal{L}_{\rm v}( x_i,y_i,x_j ) &=& \max \left \{ 0, \alpha - {\rm s}(x_i,y_i) + {\rm s}(x_j,y_i) \right \}
\end{eqnarray}

Given a dataset $\mathcal{D} = {(x_i,y_i)}_{i=1}^N$, with $N$ pairs, we optimize the following equation by stochastic gradient descent~\cite{bib:TACL:2014:Socher, bib:ICCV:2015:Zhu}.
\begin{equation}
  \mathcal{L} = \sum_{i=1}^N \sum_{j=1}^N \left ( \mathcal{L}_{\rm s}( x_i,y_i,y_j ) + \mathcal{L}_{\rm v}( x_i,y_i,x_j ) \right )
\end{equation}

\section{Extendability}
The proposed architecture has extendability of new embedding networks. The steps of the extension are straightforward. We build visual and textual networks and then merge them to form a joint embedding space. In this paper, we add embedding networks using the Inflated 3D Convolutional Neural Network (I3D) model~\cite{QuoVadis} so that the networks can capture video activities. We utilize the pre-trained RGB-I3D model to extract embedding vectors from continuous 16 frames of video. Consequently, a 1024-dimensional embedding vector ${\rm f_e}(x)$ is produced for each video.

The joint space is learnt by using ${\rm f_e}(x)$ and the textual embedding vector ${\rm g_e}(y)$, and the similarity ${\rm s_e}(x,y)$ is measured in this joint embedding space. We transform ${\rm \phi(y)}$ using $\hat{W}_e \in \mathbb{R}^{D \times 1024}$ and $\hat{b}_e \in \mathbb{R}^{1024}$. The similarity aggregation is also straightforward to be extended. Specifically, we concatenate all similarities and merge them with a weight $W'_m \in \mathbb{R}^M$, where $M$ represents a number of embedding spaces. $M=3$ in this case.
\begin{eqnarray}
  {\rm g_e}(y) &=& \hat{W}_e \phi(y) + \hat{b}_e\\
  {\rm s_e}(x,y) &=& \frac{{\rm f_e}(x) \cdot {\rm g_e}(y)}{\parallel {\rm f_e}(x) \parallel \parallel {\rm g_e}(y)\parallel}
\end{eqnarray}

We stress that the extendability is vital since this enables us to incorporate other feature extraction models into the framework quickly.
There are abundant approaches to extract features from videos~\cite{bib:ICCV:2017:Qiu, bib:CVPR:2018:Wang, bib:ECCV:2016:Wang:TSN, bib:ICCV:2015:Tran:Spatio, bib:arXiv:2018:Feichtenhofer, bib:ICLR:2020:Zhang:V4D, bib:NIPS:2014:Simonyan:two, bib:CVPR:2008:Laptev, bib:ECCV:2006:Dalal:Human, bib:ICCV:2013:Wang:Action}.
Various aspects and approaches are necessary for the understanding of videos and sentences.

\section{Experiments}
We carried out the sentence-to-video retrieval task on the benchmark dataset to evaluate the proposed method. The task is retrieving the video associated with the query sentence from the test videos. We calculated similarities over the test videos with the query sentence, and then we picked up videos according to the similarities in descending order.

We reported the experimental results using rank-based performance metrics, i.e., Recall@$k$ and Median rank. The Recall@$k$ calculates the percentage of the correct video in top-$k$ retrieved results. In this paper, we set $k= 1, 5, 10$. Median rank calculates the median of the ground-truth results in the ranking. For Recall@$k$, the bigger value indicates better performance. When Median Rank is a lower value, retrieved results are closer to the ground-truth items. Therefore, a lower median rank means better retrieval performance.

Following the convention in sentence-to-video retrieval, we used the Microsoft Research Video to Text dataset (MSR-VTT)~\cite{Xu2016MSRVTTAL}, which is a large-scale video benchmark for video understanding. The MSR-VTT contains 10000 video clips from YouTube with 41.2 hours in total. The dataset provides videos in 20 categories, e.g., music, people, and sport. Each video is associated with 20 different description sentences. The dataset consists of 6513 videos for training, 497 videos for validation, and 2990 videos for testing.

We evaluated four variants of the proposed method: single-space, sequential or I3D dual-space, and triple-space models.
Firstly, the single-space model represents the proposed framework composed of the global visual and textual embedding networks. These two networks form a single embedding space. Then, we measure the final similarity in the single embedding space. Secondly, the sequential dual-space model (dual-S) is the proposed framework using two embedding spaces learned by the global and sequential visual networks, and textual embedding networks. We measure the final similarity by merging two similarities, ${\rm s_g}$ and ${\rm s_s}$, as described in Eq.~\eqref{eq:sim:dual}. Thirdly, the I3D dual-space model (dual-I) has global visual and I3D embedding networks. Lastly, the triple-space model added the I3D and textual embedding networks into the dual-space model. We produce the final similarity by merging ${\rm s_g}$, ${\rm s_s}$, and ${\rm s_e}$ with the similarity aggregation.

\subsection{Sentence-to-Video Retrieval Results}
We summarized the results of the sentence-to-video retrieval task on the MSR-VTT dataset in Table~\ref{tab:MSRVTT}. We compared the proposed method to the existing methods~\cite{Kiros2014UnifyingVE,Bib:BMVC:2018:VSE++,Mithun2018JointEW,Dong2018DualEF} The proposed method obtained 7.1 (dual) at R@1, 21.2 (triple) at R@5, 32,4 at R@10 (triple), and 29 at MR (triple). These are competitive with~\cite{Mithun2018JointEW,Dong2018DualEF}, which are the state-of-the-art.

\begin{table}[t] \centering
  \caption{The results of sentence-to-video retrieval task on MSR-VTT dataset. The bold and underlined results represent the first- and the second-best, respectively.} \label{tab:MSRVTT}
  \begin{tabular}{l|cccc}
    Method &R@1 &R@5 &R@10 &MR\\ \hline \hline
    VSE~\cite{Kiros2014UnifyingVE} &5.0 &16.4 &24.6 &47\\
    VSE++~\cite{Bib:BMVC:2018:VSE++} &5.7 &17.1 &24.8 &65\\
    Multi-Cues~\cite{Mithun2018JointEW} &7.0 &20.9 &29.7 &38\\
    Cat-Feats~\cite{Dong2018DualEF} &\textbf{7.7} &\textbf{22} &\underline{31.8} &32\\ \hline
    Ours (single) &5.6 &18.4 &28.3 &41 \\
    Ours (dual-S) &\underline{7.1} &19.8 &31 &\underline{30}\\
    Ours (triple) &6.7 &\underline{21.2} &\textbf{32.4} &\textbf{29}
  \end{tabular}
\end{table}

VSE and VSE++ adopt a triplet ranking loss, and VSE++ incorporates hard-negative samples into the loss to facilitate practical training~\cite{bib:CVPR:2016:OEHM}. We adopted this strategy. The results show that VSE++ performs significantly better than VSE at every R@$k$. Although the single-space model and VSE++ adopted similar loss functions, we found slight improvements in performance. However, the dual-space model achieves much better performance compared to VSE++. The results demonstrate the importance of using the sequential visual information of videos for learning an embedding space.

Multi-Cues~\cite{Mithun2018JointEW} calculates two similarities in separated embedding spaces and then averages them to produce a final similarity. The proposed method has higher performance compared to Multi-Cues. The similarity aggregation strategy is the main difference between the proposed method and Multi-Cues. Note that the average suffers from aligning videos and sentences due to their variations. Thus, some videos need global visual features, and some need sequential visual features. The proposed method has a flexible strategy for merging similarities. The experimental results show that the proposed strategy is more useful for measuring similarity than a naive merging approach with equal importance to each embedding space, such as average.

The Cat-Feats~\cite{Dong2018DualEF} embeds videos and sentences by concatenating feature vectors extracted by three embedding modules: CNN, bi-directional GRU, and max pooling. Cat-Feats is slightly better than the proposed method at R@1 and R@5, e.g., 7.7 and 7.1 for Cat-Feats and dual-space at R@1, respectively. Whereas, the proposed method (triple-space) outperforms Cat-Feat at R@10 and median rank, such as 32 and 29 by Cat-Feat and the triple, respectively. These results imply that the proposed method and Cat-Feats can assist each other. There is a possibility to improve performance by incorporating the feature concatenation mechanism used in Cat-Feat into the proposed method.

Finally, the proposed method with triple-space achieves better results than single- and dual-space at three metrics: R@5, 10, and median rank. Therefore, The results show that integrating multiple similarities can lead to a better, reliable retrieval performance.

\section{Ablation Study}
We carried out ablation studies to evaluate the components of the proposed method. We conducted the following three experiments.

\subsection{Embedding Spaces}
We changed the number of embedding spaces and developed the four variants of the proposed architecture. The experimental results are shown in Table~\ref{tab:ablation:spaces}. there are certain improvements from single to multiple spaces at all the metrics. Therefore, we can verify the effectiveness of the proposed method that integrates multiple spaces. Subsequently, we compared the two duals, dual-S is better than the dual-I at R@1 and R@10, whereas dual-I is superior at R@5. Thus, dual-S and dual-I can complement each other. The triple contains the embedding spaces of both of the duals, and it outperforms the single and the duals. Therefore, we can confirm the effectiveness of the combination of multiple embedding spaces, which is the key insight of the proposed method for video and sentence understanding.

\begin{table}[t] \centering
  \caption{Evaluation on combinations of embedding spaces} \label{tab:ablation:spaces}
  \begin{tabular}{lccc|cccc}
    &\multicolumn{3}{c}{Embedding spaces} &&&& \\
    &Global & Sequential &I3D &R@1 &R@5 &R@10 &MR\\ \hline
    single &$\surd$ & & &5.6 &18.4 &28.3 &41\\
    dual-S &$\surd$ &$\surd$ & &7.1 &19.8 &31 &30\\
    dual-I &$\surd$ & &$\surd$ &6.8 &20.2 &30.6 &30\\
    triple &$\surd$ &$\surd$ &$\surd$ &6.7 &21.2 &32.4 &29
  \end{tabular}
\end{table}

\subsection{Spatial Attention Mechanism}
We conducted experiments using dual-S with or without the spatial attention mechanism.
The dual-S without the attention encodes each frame using ResNet into a 2048-dimensional vector. Then, the LSTM processed the sequences of the vectors. Table~\ref{tab:ablation:attention} shows the results. The dual-S with the attention achieves better results than without attention at all metrics, R@1, R@5, R@10, and median rank. Therefore, we can confirm that the proposed spatial attention mechanism improves performance significantly.

Besides, we observed that the performance of the dual-S with attention is almost competitive with dual-I. However, dual-S without the attention is worse than the dual-I. Considering that the I3D model extracts useful spatial and temporal features for action recognition, the dual-S without attention could not extract sequential features effectively. Whereas, the dual-S with attention obtained such features. We stress that this is another supportive evidence showing the effectiveness of the proposed attention mechanism.

\begin{table}[t] \centering
  \caption{Effectiveness of the spatial attention} \label{tab:ablation:attention}
  \begin{tabular}{l|cccc}
    &R@1 &R@5 &R@10 &MR \\ \hline
    w/o attention &5.8 &19.8 &28.6 &34 \\
    w/ attention &7.1 &19.8 &31 &30
  \end{tabular}
\end{table}

\subsection{Similarity Aggregation}
We investigated the impacts of similarity aggregation using average or the proposed weighted sum. We used dual-S and dual-I for this investigating experiment. The experimental results are shown in Table~\ref{tab:ablation:agg}. There are improvements by the weighted sum at R@1, R@10, and MR in both of the dual-S and dual-I. Therefore, we confirmed the effectiveness of the proposed similarity aggregation module.

\begin{table}[t] \centering
  \caption{Performance comparison on similarity aggregation using average or the proposed weighted sum} \label{tab:ablation:agg}
  \begin{tabular}{ll|cccc}
    & &R@1 &R@5 &R@10 &MR \\ \hline \hline
    \multirow{2}{*}{dual-S} &average &6.6 &19.9 &29.9 &31\\
    &weighted &7.1 &19.9 &31 &30\\ \hline
    \multirow{2}{*}{dual-I} &average &6.7 &20.2 &30.4 &30 \\
    &weighted &6.8 &20.2 &30.6 &30
  \end{tabular}
\end{table}

Furthermore, we performed an analysis of the weights in the similarity aggregation. As described in Eq.~\eqref{eq:aggregation:weight}, the weights are flexibly determined according to the given sentence. In other words, the weight represents the importance of embedding spaces. We attempted to go further understanding by analyzing the weights. For simplicity, we used the dual-S model in this analysis. Therefore, the analysis is on the importance of global and sequential visual features. We summarized the statistics of the weights in Table~\ref{tab:ablation:analysis}. The statistics show that the proposed method assigns larger weights to the global features.

\begin{table}[t] \centering
  \caption{Statistics of the weights in similarity aggregation for the global and sequential embedding spaces} \label{tab:ablation:analysis}
  \begin{tabular}{l|ccc}
    &Average &Min &Max\\ \hline
    Global &0.52 &0.399 &0.61\\
    Sequential &0.48 &0.393 &0.60
  \end{tabular}
\end{table}

\begin{figure}[!t] \centering
  \includegraphics[width=3.5in]{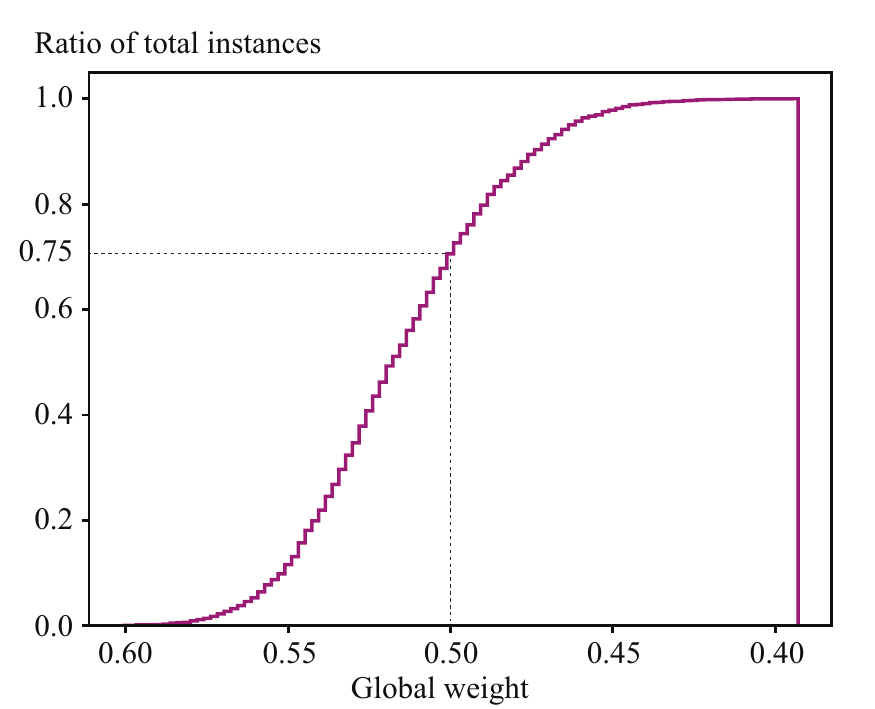}
  \caption{Accumulative step histogram of the weights of the global visual features used in similarity aggregation} \label{fig:ablation:histogram}
\end{figure}

We show the accumulative step histogram for the global weight in Fig.~\ref{fig:ablation:histogram}. The ratio reached 0.75 at the weight 0.5. Thus, 0.75 total instances received larger weights for the global feature. In contrast, the weights of the sequential feature are larger only in the remained 0.25 instances. Therefore, the global feature is more critical than the sequential feature. Thus, dual-S aggressively used global features.

\begin{figure*}[!t] \centering
  \includegraphics[width=7.2in]{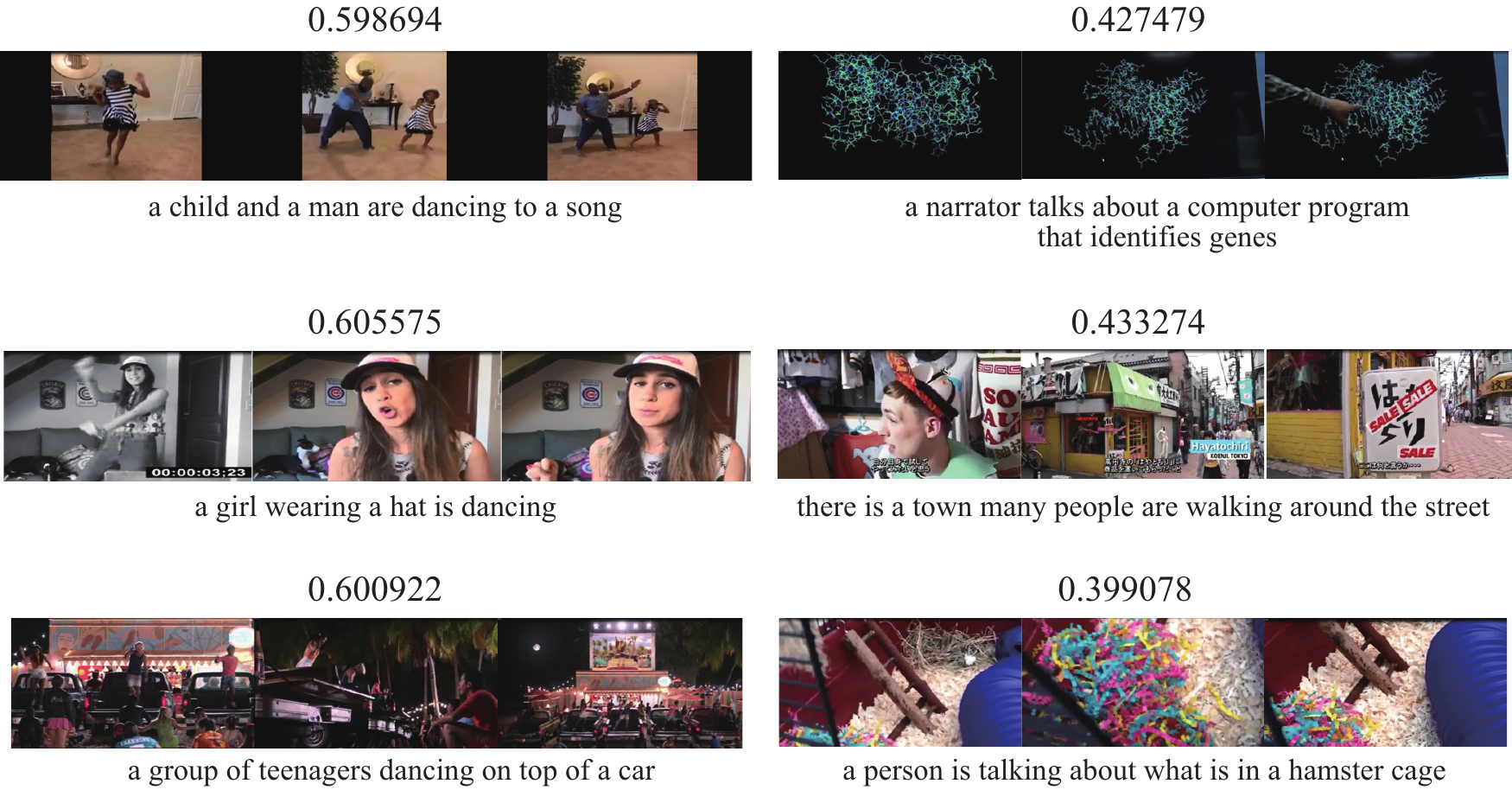}
  \caption{Examples of video and sentence with assigned weight to the global feature. The numbers and the sentences represent the weight for the global feature and query sentences, respectively.} \label{fig:ablation:examples}
\end{figure*}

Figure~\ref{fig:ablation:examples} shows examples of videos and sentences with assigned weight to the global feature. Videos containing explicit scenes tend to have larger weights on the global visual feature since objects in the videos are relatively easy for detection. On the other hand, videos containing unclear scenes tend to assign larger weights to the sequential visual feature.

%

\section{Conclusion}
In this paper, we presented a novel framework for embedding videos and sentences into multiple embedding spaces. The proposed method uses distinct embedding networks to capture various relationships between visual and textual features, such as global appearance, sequential visual, and action features. We produce the final similarity between a video and a sentence by merging similarities measured in the embedding spaces with the weighted sum manner. The proposed method can flexibly determine the weights according to a given sentence. Hence, the final similarity can incorporate an essential relationship between video and sentence.

We carried out sentence-to-video retrieval experiments on the MSR-VTT dataset to demonstrate that the proposed framework significantly improved the performance when the number of embedding spaces increased. The proposed method achieved competitive results compared to the state-of-the-art methods~\cite{Mithun2018JointEW, Dong2018DualEF}. Furthermore, we verify all the critical components in the proposed method through the ablation experiments. Even though the components are individually useful, their cooperation can generate significant improvements.

\section*{Acknowledgment}
This study was partially supported by JSPS KAKENHI Grant Number 18K19772, 19K11848, and Yotta Informatics Project by MEXT, Japan.

\bibliographystyle{IEEEtran}
\bibliography{refs}

\newpage

\end{document}